\documentclass{article}

\usepackage{microtype}
\usepackage{graphicx}
\usepackage{subfigure}
\usepackage{booktabs} % for professional tables
\usepackage{amsmath,amsfonts,amssymb,mathtools}
\usepackage{hyperref}
\allowdisplaybreaks
\usepackage[accepted]{icml2021}

\newcommand{\etal}{\textit{et al}.}
\newcommand{\ie}{\textit{i}.\textit{e}., }

\newcommand{\algoname}{\text{Master-User}}
\newcommand{\archi}{\text{Col-NN}}

\icmltitlerunning{Columnar Networks for Recurrent Learning}

\begin{document}

\twocolumn[
\icmltitle{Scalable Online Recurrent Learning Using \\ Columnar Neural Networks}



\icmlsetsymbol{equal}{*}

\begin{icmlauthorlist}
\icmlauthor{Khurram Javed}{to}
\icmlauthor{Martha White}{to}
\icmlauthor{Rich Sutton}{to,dmorg}
\end{icmlauthorlist}

\icmlaffiliation{to}{RLAI Lab, University of Alberta, Edmonton}
\icmlaffiliation{dmorg}{DeepMind, Edmonton}

\icmlcorrespondingauthor{Khurram Javed}{kjaved@ualberta.ca}

\icmlkeywords{Machine Learning, ICML}

\vskip 0.3in
]



\printAffiliationsAndNotice{\icmlEqualContribution} % otherwise use the standard text.

\begin{abstract}
% Learning in neural networks requires structural credit-assignment --- identifying parameters that influence a prediction made by the network. For recurrent learning, a parameter can influence a prediction made many steps in the future making credit-assignment challenging. Two popular gradient-based algorithms for structural credit-assignment are (1) Back-propagation Through Time (BPTT) and (2) Real-time Recurrent Learning (RTRL). BPTT requires memory proportional to the length of the input sequence and scales poorly.
% % Additionally, it does not spread the operations for gradient computation uniformly across time. 
% RTRL, on the other hand, can compute gradients in real-time for arbitrarily long sequences using constant memory but is computationally intractable for large networks. In this work, we propose a network architecture --- Columnar Neural Networks (\archi{}) --- and a credit-assignment algorithm --- \algoname --- that allow us to approximate gradients in real-time using $O(n)$ operations and memory per-step. Our method builds on the idea that for modular recurrent networks composed of columns with scalar states it is sufficient for a parameter to track its influence on the state of its own column. As long as connections across columns are sparse, our approximation is close to the true gradient. \archi{} trained with \algoname can be applied for learning recurrent states and for meta-learning using two traces per-parameter.

Structural credit assignment for recurrent learning is challenging. An algorithm called RTRL can compute gradients for recurrent networks online but is computationally intractable for large networks. Alternatives, such as BPTT, are not online. In this work, we propose a credit-assignment algorithm --- \algoname{} --- that approximates the gradients for recurrent learning in real-time using $O(n)$ operations and memory per-step.  Our method builds on the idea that for modular recurrent networks, composed of columns with scalar states, it is sufficient for a parameter to only track its influence on the state of its column. We empirically show that as long as connections between columns are sparse, our method approximates the true gradient well. In the special case when there are no connections between columns, the $O(n)$ gradient estimate is exact. We demonstrate the utility of the approach for both recurrent state learning and meta-learning by comparing the estimated gradient to the true gradient on a synthetic test-bed. 
\end{abstract}

\section{Introduction}
Structural credit-assignment --- identifying how to change network parameters to improve predictions --- is an essential component of learning in neural networks. Effective structural credit-assignment requires tracking the influence of parameters on future predictions. A parameter can influence a prediction in the future in two primary ways. First, for recurrent networks (RNNs), a parameter can influence the internal state of the network which, in turn, can affect a prediction made many steps in the future. Second, if the network is learning online, a parameter can influence the learning updates. These learning updates, in turn, influence predictions made in the future. Structural credit-assignment through recurrent states is called recurrent state learning, whereas through the learning updates is called meta-learning (Schmidhuber,~1987; Bengio~\etal,~1990; and Sutton,~1992). 

Back-Propagation Through Time (BPTT)~(Werbos, 1988; Robinson and Fallside, 1987) is a popular algorithm for gradient-based structural credit-assignment in RNNs. BPTT extends the back-propagation algorithm for feed-forward networks --- independently proposed by Werbos~(1974) and Rumelhart~\etal~(1986) --- to RNNs by storing network activations from prior steps, and repeatedly applying the chain-rule starting from the output of the network and ending at the activations at the beginning of the sequence. BPTT is unsuitable for online learning as it requires memory proportional to the length of the sequence. Moreover, it delays gradient computation until the end of the sequence. For online learning, this sequence can be never-ending or arbitrarily long.

\begin{figure*}
\centering
  \includegraphics[width=0.98\textwidth]{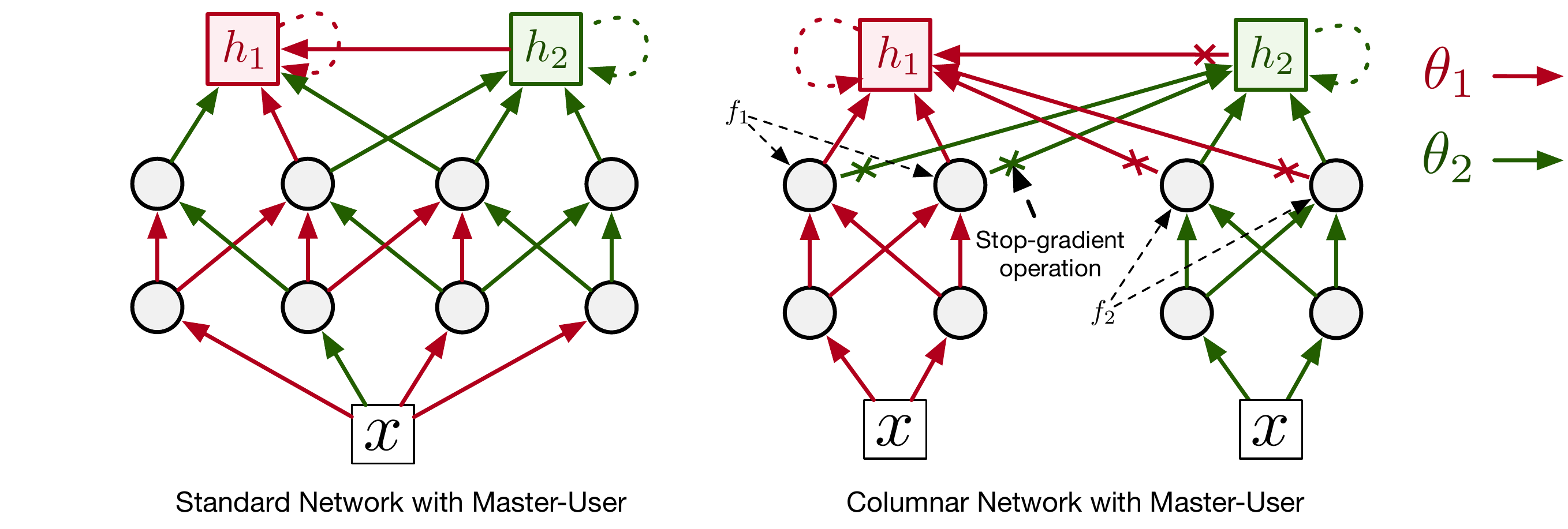}
  \caption{We can divide $\theta$ into $\{\theta_1, \theta_2\}$ randomly, or in columns. The red parameters are $\theta_1$ and green $\theta_2$. Our goal is to estimate $\frac{\partial h_i}{\partial \theta_i}$ for $i=1$ and $2$. For the network on the left, this would require two backward passes --- one starting from $h_1$ and one from $h_2$. For the Columnar network on the right, on the other hand, we can back-propagate gradients starting from the sum $h_1 + h_2$ and stop the flow of gradients from parameters connecting the features from one column to the feature of others. We show the stop-gradient operations with '$\times$' on the arrows. The stop-gradient operations allows us to compute $\frac{\partial h_i}{\partial \theta_i}$ exactly for $i=1$ and $2$ in a single backward pass. Organizing parameters in columns is key to estimating the required gradient in a single pass.}
  \label{lateral_connections}
\end{figure*}

% The dependence of BPTT on storing activations from past hinders its application to online learning. Additionally, the computation in BPTT is not spread uniformly across time --- the learner accumulates activations for a sequence in the memory and delays the computation of the gradients until the end. This is incompatible with the goal of real-time learning --- the ability to incorporate feedback from the environment quickly for correcting mistakes. 

RTRL --- an alternative to BPTT --- was proposed by Williams and Zipser~(1989). RTRL relies on forward-mode differentiation --- using chain-rule to compute gradients in the direction of time --- to compute gradients recursively. Unlike BPTT, RTRL does not delay gradient-computation until the final step. The memory requirement of RTRL also does not depend on the sequence length. As a result, it is a better candidate for real-time online learning. Unfortunately, RTRL requires maintaining the Jacobian $\frac{\partial h(t)}{\partial \theta}$ at every step, which requires $O(|h||\theta|)$ memory, where $|h|$ is the size of state of the network and $|\theta|$ is the number of total parameters. The Jacobian is recursively updated by applying chain rule as:  
$$\frac{\partial h(t+1)}{\partial \theta} =\frac{\partial h(t+1)}{\partial \theta(t+1)} +  \frac{\partial h(t+1)}{\partial h(t)}\frac{\partial h(t)}{\partial \theta}, $$ 
which requires  $O(|h|^2|\theta|)$ operations and scales poorly to large networks.

% For a more thorough explanation of the memory and time complexity of recurrent state-learning and meta-learning methods using RTRL, see Appendix A. 

A promising direction to scale gradient-based credit-assignment to large networks is to approximate the gradient. Elman~(1990) proposed to ignore the influence of parameters on future predictions completely for training RNNs. This resulted in a scalable but biased algorithm. Williams and Peng~(1990) proposed a more general algorithm called Truncated BPTT (T-BPTT). T-BPTT tracks the influence of all parameters on predictions made up to $k$ steps in the future. T-BPTT is implemented by limiting the BPTT computation to last $k$ activations and works well for a range of problems~(Mikolov~\etal, 2009, 2010; Sutskever, 2013 and Kapturowski~\etal, 2018). Its main limitation is that the resultant gradient is blind to long-range dependencies. Mujika~\etal~(2018) showed that on a simple copy task, T-BPTT failed to learn dependencies beyond the truncation window. Tallec~\etal~(2017) demonstrated T-BPTT can even diverge when a parameter has a negative long-term effect on a target and a positive short-term effect.

RTRL can also be approximated to reduce its computational overhead. Ollivier~\etal~(2015) and Tallec~\etal~(2017) proposed NoBacktrack and UORO. Both of these algorithms provide stochastic unbiased estimates of the gradient and scale well. However, their estimates have high variance and require extremely small step sizes for effective learning. Cooijmans and Martens~(2019) and Menick~\etal~(2021) showed that, for practical problems, UORO does not perform well due to its high variance compared to other biased approximations. Menick~\etal~(2021) proposed an approximation to RTRL called SnAp-$k$. SnAp-$k$ tracks the influence of a parameter on a state only if the parameter can influence the state within $k$ steps. It first identifies parameters whose influence on a state is zero for $k$ steps and then assumes the future influence to be zero as well. For the remaining parameters, it tracks their influence on all future predictions. T-BPTT, on the other hand, ignores the influence of all parameters on all predictions made after $k$ steps. This makes SnAp-k less biased than T-BPTT with a truncation window of $k$. SnAp-1 can be computationally efficient but introduces significant bias. SnAp-$k$ for $k>1$ reduces bias but can be as expensive as RTRL for dense RNNs.  Menick~\etal~(2021) further proposed using sparse connections as a way to make SnAp more scalable. Connection sparsity reduces the number of parameters that can influence a state within $k$ steps
%  SnAp, combined with sparsity, is a promising research direction.
 Menick~\etal~(2021) showed that for highly sparse networks, SnAp-$k$ reduces the computational requirement of RTRL by over 95\% while keeping bias in check. One limitation of their work is that they do not provide a scalable method to identify parameters that would not influence a state within $k$ steps. Instead, they run the RNN for $k$ steps and look at the full Jacobian $\frac{\partial h(t)}{\partial \theta}$ to determine these parameters. For large networks, computing this Jacobian even once is not possible. Moreover, because they induce sparsity randomly in their networks, the resultant sparsity in the Jacobian is not structured and is not amenable to efficiency gains using existing hardware. Finally, SnAp does not scale well when used in conjunction with deep feature extractors. If internal states operate on a shared representation computed with a deep network with parameters $\theta$, even SnAp-1 could require $O(|\theta||h|)$ memory and $O(|\theta||h|^2)$ operations per-step. Our goal is to design an algorithm that requires $O(|\theta|)$ memory and operations.

We propose an algorithm --- called \algoname{} --- and a recurrent network architecture --- called Columnar Neural Networks (\archi s). \algoname{} requires $O(|h||\theta|)$ operations and $O(|\theta|)$ memory per-step for credit assignment in existing recurrent networks. When combined with \archi s, the time complexity of \algoname{} reduces to $O(|\theta|)$. The central idea behind \archi s is that recurrent learning can be made more efficient by modularizing the network. For a special class of \archi s, \algoname{} can even estimate the exact gradients in $O(|\theta|)$ time.

% \begin{figure*}
% \centering
%   \includegraphics[width=0.8\textwidth]{columnar_networks_paper/figures/gradient_accuracy.pdf}
%   \caption{Accuracy of the gradient-estimate compared to the true gradient computed using BPTT. x-axis shows the normalized number of lateral connections between columns, where 1 means maximum possible lateral connections and 0 means no lateral connections. Y-axis shows the quality of approximation. The left side plot shows percentage of parameters of which the estimated gradient points in the same direction as the true gradient whereas the plot on the right shows the absolute error in the gradient estimate. Note that the left plot using the log scale for the x-axis. When no of lateral connections is low, the estimate of the gradient is good. As we move towards fully connected networks and lose the structure of the columnar networks, the gradient approximation gets bad. For fully connected networks, 40\% of the parameters have gradients pointing in the opposite direction. This is close to random gradient, which would point in random directions 50\% of the time.}
%   \label{results_rnn}
% \end{figure*}

\section{Problem Formulation}
Let $\theta \in \mathbb R^{p} $ be the parameters of a recurrent network and $h(t)\in~\mathbb R^n$ be the hidden state at time $t$. The network combines the state $h(t)$ linearly using $w(t)\in \mathbb R^n $ to make a prediction $y(t)$ as:
\begin{equation}
y(t) = \sum_{i=1}^{n} w_i(t)h_i(t) \label{pln}
\end{equation}

Let $y^*(t)$ be the target at time $t$. Then the error in the prediction is given by:

\begin{equation*}
    \delta(t) = y^*(t) - y(t).
\end{equation*}
Our goal is to minimize the error 
\begin{equation*}
    \mathcal L(t) = \frac{1}{2}\delta^2(t).
\end{equation*}
The goal of a gradient-based credit-assignment algorithm can be formalized as estimating or approximating credit given by:
\begin{align}
G(t) &= \frac{\partial \mathcal L(t)}{\partial \theta} \label{credit}
% &= \frac{\partial \delta^2(t)}{2 \partial \theta} \\ 
% &= \delta(t)\frac{\partial \delta(t)}{\partial \theta}  \\ 
% &= \delta(t)\frac{\partial (y^*(t) - \sum_{i=1}^{n} w_i(t)h_i(t))}{\partial \theta} \\
% &= -\delta(t)\frac{\partial (\sum_{i=1}^{n} w_i(t)h_i(t))}{\partial \theta} 
\end{align}
Here $\theta$ is not indexed by $t$ because we want to estimate the impact of $\theta$ at all prior time-steps on the current prediction. Since computing this exact gradient using BPTT or RTRL is expensive, we aim to approximate it. We propose an algorithm called \algoname.  
% Let $C(t) \in  \mathbb R^{p}$ be the credit assigned to each state. Then, $C(t)$ is a useful approximation of $G(t)$ if they are aligned --- they have the same sign. 
 
 \section{The \algoname{} Algorithm}
 \label{section3}
 The central idea behind \algoname{} is to divide the parameters of an RNN into $n$ disjoint groups --- $\theta = \{\theta_1, \theta_2, \cdots, \theta_n\}$ --- and assign every group to a scalar hidden state. Let  $\theta_i$ be assigned to $h_i$. Then, \algoname{} estimates the influence of  $\theta_i$ on $h_i(t)$ by approximating $\frac{\partial h_i(t)}{\partial \theta_i}$ online. For $i\ne j$, it ignores the influence of $\theta_i$ on $h_j(t)$ and treats $\frac{\partial h_j(t)}{\partial \theta_i}=0$ . We call $h_i$ the \textit{master state} for $\theta_i$ --- it can pass gradients to $\theta_i$ for assigning credit --- whereas $h_j$ is the \textit{user state} --- it can only use features generated by $\theta_i$, but cannot change those features by assigning credit. To keep track of influence of $\theta_i$ on $h_i$, The trace of the gradient $\frac{\partial h_i(t)}{\partial \theta_i}$ can be computed recursively as: 
 \begin{equation}
 \frac{\partial h_i(t)}{\partial \theta_i} =  \frac{\partial h_i(t)}{\partial \theta_i(t)} + \frac{\partial h_i(t)}{\partial h(t-1)} \frac{\partial h(t-1)}{\partial \theta_i}.
 \label{master_user_actual}
 \end{equation}
  The update involves $\frac{\partial h(t-1)}{\partial \theta_i}$ --- a Jacobian of size $|h|\times |\theta_i|$. Computing this Jacobian is expensive so \algoname{} approximates the update by ignoring the influence of $\theta_i$ on $h_i$ through states $h_j$ as:
  \begin{equation}
 \frac{\partial h_i(t)}{\partial \theta_i} \approx  \frac{\partial h_i(t)}{\partial \theta_i(t)} + \frac{\partial h_i(t)}{\partial h_i(t-1)} \frac{\partial h_i(t-1)}{\partial \theta_i},
 \label{master_user}
 \end{equation}
 
 where $\frac{\partial h_i(t)}{\partial \theta_i(t)}$ is the gradient of $h_i(t)$ that takes into account the influence of $\theta_i$ only at time $t$.
The terms $\frac{\partial h_i(t)}{\partial \theta_i(t)}$ and $ \frac{\partial h_i(t)}{\partial h_i(t-1)}$ can be computed in $O(|\theta|)$\footnote{Time complexity of back-propagation is often written as $O(|\theta| + k)$, where k is total number of nodes. However, $k < |\theta|$ for neural networks resulting in the bound $O(2|\theta|)=O(|\theta|)$} operations using the back-propagation algorithm . The product $\frac{\partial h_i(t)}{\partial h_i(t-1)} \frac{\partial h_i(t-1)}{\partial \theta_i}$ and the summation in Equation~\ref{master_user} take $O(|\theta_i|)$ operations. Equation~\ref{master_user} has to be repeated for each hidden state, making the overall complexity $O(|h||\theta|)$ which falls short of our goal of an order $O(|\theta|)$ algorithm.

Applying \algoname{} to standard RNNs is unlikely to work well. \algoname{} completely ignores the influence of $\theta_i$ on $h_j$ and only approximates the influence of $\theta_i$ on $h_i$. In the next section, we introduce an architecture called Columnar Neural Networks (\archi s) for which the approximations made by \algoname{} are sensible. Moreover, \algoname{} can exploit the structure in \archi s to scale linearly with number of parameters. 

% Additionally, assuming $\frac{\partial h_j(t)}{\partial \theta_i}$ to be zero for existing dense RNN architecture, such as LSTM~(Hochreiter, 1997) or GRU~(Cho, 2014) is unjustified and the resulting estimate of the gradients is unlikely to be a good approximation. Consequently, \algoname{} alone does not satisfy the goal of scaling linearly with the number of parameters and is unlikely to approximate the true gradient. 

 \begin{figure}
\centering
  \includegraphics[width=0.48 \textwidth]{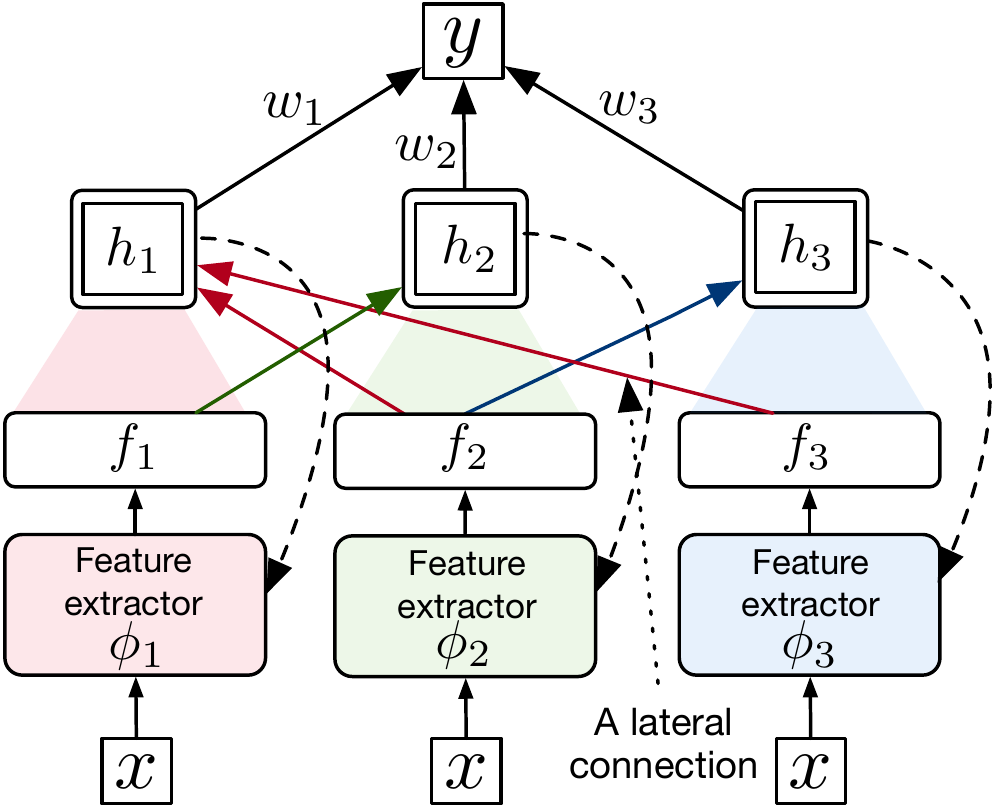}
%   \caption{Architecture for Columnar Neural Network (\archi{}). A \archi{} has an n-dimensional hidden state --- one outputted by each column. The parameters in column $i$ are $\theta_i$ and they only keep track of their influence on $h_i$ over time. Column $j$ feature extractor is free to use features from column $i$ using lateral connections. However, $\theta_i$ ignores how they influence $h_j$. Increasing the size of a \archi{} by adding more columns or increasing the size of each column only linearly increases the memory and operations per-step for structural credit-assignment.}
  \caption{A three column \archi{}. The red region and red arrows represent the parameters for the first column. Arrows connecting a column with the state of a different column --- such as the red arrow from $f_3$ to $h_1$ --- are called lateral connections. Lateral connections allow a column to use features learned by other columns, but they also introduce error in the estimate of the gradient. When there are no lateral connections, exact gradient for a \archi{} can be computed in $O(|\theta|)$ time.}
  \label{cnn}
\end{figure}

 \section{Columnar Neural Networks}
 \algoname{} divides the parameter into $n$ disjoint groups, but does not specify how these groups are constructed. Parameters in $\theta_i$ can be spread all over the network, or assigned randomly. Columnar Neural Networks (\archi s) impose further structure on these groups by organizing them into $n$ columns. Every column has an associated feature extractor. The feature extractor for the $i$th column, $C_{\phi_i}$, is parameterized by $\phi_i$. It takes as input $x(t)$ and $h_i(t-1)$ and outputs a feature vector $f_i(t) \in \mathbb R^k$ \ie
 \begin{equation*}
     f_i(t) = C_{\phi_i}(x(t), h_i(t-1))
 \end{equation*}
Hidden state $h_i$ is updated as: 
 \begin{equation*}
     h_i(t) = \mathcal R_{u_i}(f_1(t), f_2(t), \cdots, f_n(t),  h(t-1))
 \end{equation*}
 where $\mathcal R_{u_i}$ is parameterized by $u_i$ and defines a recursive relationship between $h_i(t)$ and $h_i(t-1)$. The set of parameters $\{u_i, \phi_i\}$ constitute the group $\theta_i$. $\mathcal R_{u_i}$ can be used to implement existing recurrent architectures, such as an LSTM or GRU cell. An abstracted view of a \archi{} is shown in Figure~\ref{cnn}. 
 
 \archi{} restricts the architecture of the feature extractor, but provides a key benefit from an optimization point of view --- it can reduce the overhead of \algoname{} from $O(|h||\theta|)$ to $O(|\theta|)$. This is possible because for a \archi{}, the set of gradients $\{\frac{\partial h_i(t)}{\partial \theta_i(t)}\}_{i=1}^n$ --- required by Equation~\ref{master_user} --- can be computed by using back-propagation only once.
 
 \subsection{$O(|\theta|)$ Implementation of \algoname{} for \archi s}
 \label{efficient_implementation}
 In Section~\ref{section3}, we noted that we need a separate application of back-propagation for computing $\frac{\partial h_i(t}{\partial \theta_i(t)}$ for each $i$. For \archi s, it is possible to compute $\frac{\partial h_i(t}{\partial \theta_i(t)}$  for all $i$ with a single application of back-propagation. 
 
 We modify back-propagation by restricting the backward flow of gradient from state $h_j$ to features $f_i$, as shown in Figure~\ref{lateral_connections}. We then compute the gradient of the sum of the states $\frac{\partial (\sum_{k=1}^n h_k(t))}{\partial \theta(t)}$; the resultant gradient is $\{\frac{\partial h_i(t)}{\partial \theta_i(t)}\}_{i=1}^n$. To see why, note that the gradient of the sum is equal to sum of gradients \ie

\begin{align}
    \frac{\partial \left(\sum_{k=1}^n h_k(t) \right)}{\partial \theta(t)} &=  \sum_{k=1}^n\frac{\partial h_k(t)}{\partial \theta(t)}. \label{sum}
\end{align}
If we can remove gradients $\frac{\partial h_j(t)}{\partial \theta_i(t)}$ for $j \ne i$ in Equation~\ref{sum} from the computation graph, back-propagation w.r.t the sum will give us $\frac{\partial (\sum_{k=1}^n h_k(t))}{\partial \theta(t)}$. To remove the gradients $\frac{\partial h_j(t)}{\partial \theta_i(t)}$ for $j \ne i$, we modify back-propagation so that it ignores the flow of gradient from $f_j$ to $\phi_i$ as shown in Figure~\ref{lateral_connections}~(Right). Because $h_j(t)$ can only be influenced by $\theta_i(t)$ through $f_i(t)$, ignoring gradient from $h_j(t)$ to $f_i(t)$ is all we need to ignore $\frac{\partial h_j(t)}{\partial \theta_i(t)}$ for $j \ne i$. Applying this modified back-propagation algorithm gives us $\{\frac{\partial h_i(t)}{\partial \theta_i(t)}\}_{i=1}^n$ for all parameters in the same number of operations as a single application of back-propagation. The same reasoning can be used to show that  $\{\frac{\partial h_i(t)}{\partial h_i(t-1)}\}_{i=1}^n$ can be computed in a single backward pass. These two terms, when plugged into Equation~\ref{master_user},  give us $\{\frac{\partial h_i(t)}{\partial \theta_i}\}_{i=1}^n$ in $O(|\theta|)$ operations.

\section{Credit-Assignment in Recurrent Learning}
\algoname{} provides the gradient estimate for all parameters w.r.t their master state. We use this estimate to approximate $G(t)$ introduced in Equation~\ref{credit} as follows: 

\begin{align}
G(t) &= \frac{\partial \mathcal L(t)}{\partial \theta_i} \notag \\
&= \frac{\partial \delta^2(t)}{2 \partial \theta_i} \notag \\ 
&= \delta(t)\frac{\partial \delta(t)}{\partial \theta_i} \notag  \\ 
&= \delta(t)\frac{\partial (y^*(t) - \sum_{j=1}^{n} w_j(t)h_j(t))}{\partial \theta_i} \notag \\
&= -\delta(t) \sum_{j=1}^{n} w_j(t) \frac{\partial  h_j(t)}{\partial \theta_i} \label{before_rnn}  \\
& \text{Assuming $\frac{\partial  h_j(t)}{\partial \theta_i}=0$ when $i\ne j$} \notag  \\
&\approx  -\delta(t) w_i(t) \frac{\partial  h_i(t)}{\partial \theta_i}. \label{final_credit_rnn}
\end{align}
All terms in Equation~\ref{final_credit_rnn} can be computed in $O(|\theta|)$ time and memory for \archi, as discussed in the previous section. 

\subsection{Making Sense of the Approximations}
Credit-assignment using \algoname{} required two approximations. First, we approximated: 
\begin{equation}
    \frac{\partial h_i(t)}{\partial h_j(t-1)} \frac{\partial h_j(t-1)}{\partial \theta_i} \approx 0  \text{ when } i \ne j \label{irre}
\end{equation}
to simplify Equation~\ref{master_user_actual} to Equation~\ref{master_user}. Second, we approximated: 

\begin{equation}
    \frac{\partial  h_j(t)}{\partial \theta_i} \approx 0 \text{ when } i \ne j \label{approx_all_you_need}
\end{equation}
to simplify Equation~\ref{before_rnn} to Equation~\ref{final_credit_rnn}. If the approximation in Equation~\ref{approx_all_you_need} holds, then the approximation in Equation~\ref{irre} follows.

A \archi{} provides an easy way to make $ \frac{\partial  h_j(t)}{\partial \theta_i} \approx 0$. In a \archi{}, $\theta_i$ can only influence $h_j(t)$ through the feature $f_i(t)$. Let the connections from $f_i(t)$ to $h_j(t)$ for $i \ne j$ be called \emph{lateral connections} as shown in Figure~\ref{cnn}. Then, removing all lateral connections would make the approximation in Equation~\ref{approx_all_you_need} to hold exactly, allowing us to compute the true gradient in $O(|\theta|)$ time. A downside of removing all lateral connections is that $h_i(t)$ can no longer use features generated by $\theta_j$, or information stored in $h_j(t-1)$, to update itself. Instead of removing lateral connections entirely, we propose to make lateral connections sparse to reduce the influence of $\theta_i$ on $h_j$. 

\subsection{Evaluating Quality of the Gradient Estimate}
\label{align_exp}
\begin{figure}
\centering
  \includegraphics[width=0.45 \textwidth]{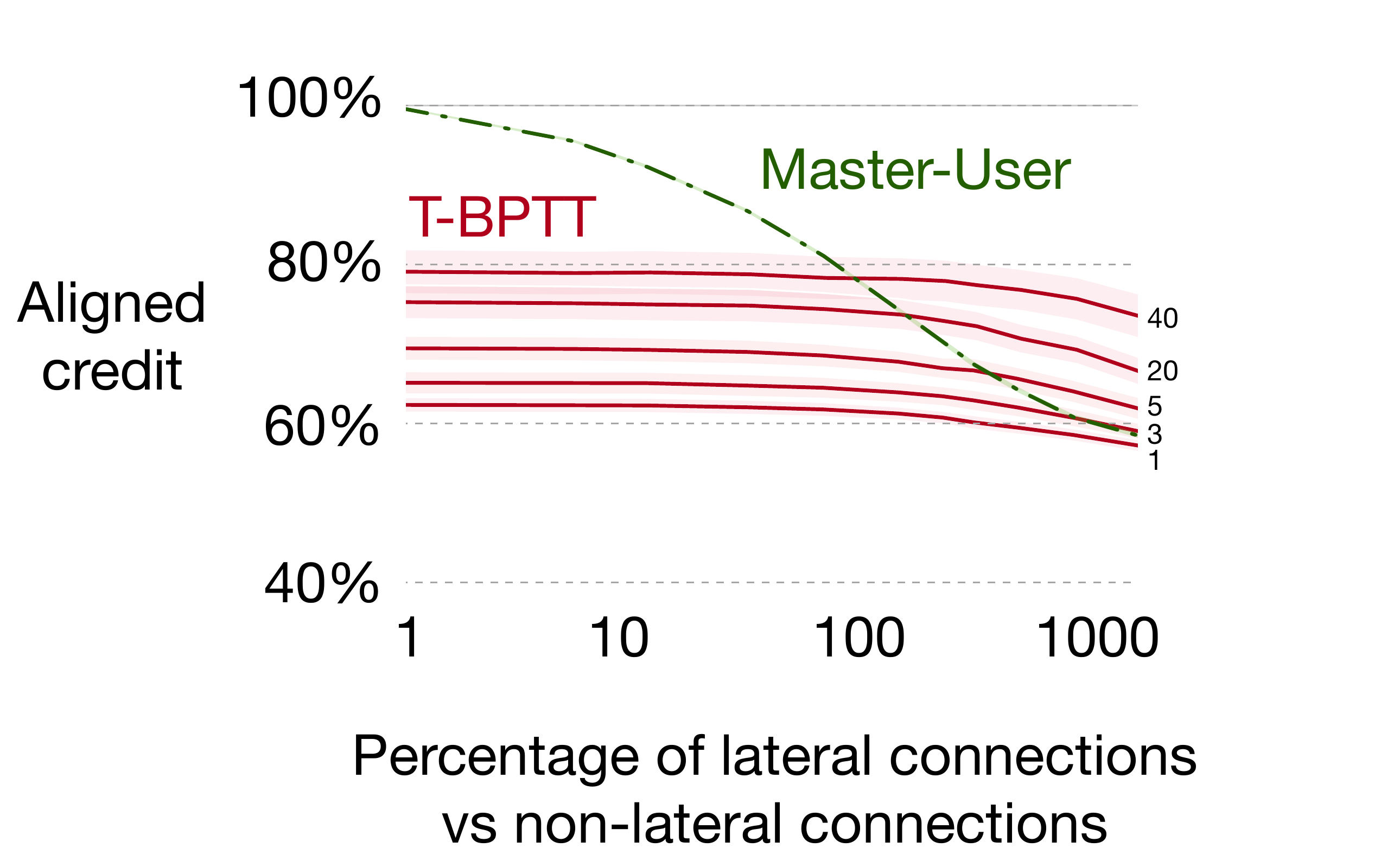}
  \caption{Quality of the approximation as a function of lateral connectivity. When number of lateral connections are low compared to non-lateral connections, the gradient estimate of \algoname{} is highly aligned with the true gradient. As lateral connections increase, the approximation gets worse. T-BPTT's estimate has significantly higher error for \archi s with sparse lateral connections even when the truncation window --- 40 --- is almost equal to the sequence length --- 50. All curves are averaged over 100 random seems, and the confidence intervals are one standard error.}
  \label{rnn_alignment}
\end{figure}

We compare the $O(n)$ gradient estimate of \algoname{} in a \archi{} to the true gradient computed using BPTT on a synthetic benchmark. 
\subsubsection{Synthetic Benchmark}
We randomly generate input and target sequences. Input $x(t)$ is a vector of length fifty. Each element of the input vector is uniformly sampled from the set $\{0, 1\}$. Target $y^*(t)$ is uniformly sampled from the range $(-50, 50)$. Random targets result in high error and large gradients, which is useful for testing the quality of the gradient estimate. 
\subsubsection{Network Architecture} Each column of the \archi{} implements a 2-layer fully connected network with 50 features in each layer and ReLU activations~(Glorot~\etal, 2011). The output of each column is a feature of length 50.  We use a total of 20 columns. The parameters of our network are initialized using Xavier Initialization~(Glorot and Bengio,~2010) to ensure a high flow of gradient. The recurrence function, $h_j(t) = \mathcal R_{u_j}(f_1(t), \cdots, f_n(t), h_j(t-1))$ is defined as:
\begin{equation}
    h_i(t) = h_i(t-1) + \sigma_g (\sum_{j=1}^n U^T_{ij} f_j(t) + R^T_i h(t-1)) \label{recurrent_used}
\end{equation}
where $U_i$ is a matrix with dimensions $|h|\times |f_j|$, $R_i$ is a vector of length $|h|$ and $\sigma_g$ is the $tanh$ function. We use recurrence defined in Equation~\ref{recurrent_used} instead of popular architectures, like LSTM~(Hochreiter~and~Schmidhuber, 1997) or GRU~(Cho~\etal, 2014), because untrained LSTM or GRU networks initialized using popular initialization techniques decay the information in recurrent cell rapidly, removing long-term dependencies. Equation~\ref{recurrent_used}, on the other hand, does not decay the old state at all resulting in long-term dependencies between parameters and predictions. Our recurrent function is a poor choice for practical problems, but is useful for evaluating the accuracy of credit-assignment algorithms for capturing long-term dependencies. For more details on the decay rate in untrained LSTM and GRU networks, and results with GRU networks, see Appendix~A. 
\subsubsection{Controlling Sparsity in Lateral Connections}
We make the weight vector $U_{ij}$ for $ i \ne j$ sparse by masking weights at random to be zero. This reduces the influence of $f_j$ on $h_i(t)$ for . Let $m$ be the number of unmasked weights in $U_{ij}$  for all $ j \ne i$ and $n$ be the number of weights in $U_{ii}$ that connect $f_i(t)$ to $h_i(t)$. Then, percentage of lateral connections vs non-lateral connections is defined as $s=\frac{m}{n}$.  When $s=100\%$, the number of lateral connections is the same as non-lateral connections whereas $s=1000\%$ implies there are ten times more lateral connections than non-lateral connections. For a visual depiction of lateral connections, see Figure~\ref{cnn}.
\subsubsection{Results}
We evaluate the accuracy of our gradient estimate by comparing it to the true gradient $\sum_{i=1}^{50} \frac{\partial \mathcal L(i)}{\partial \theta}$ and report the results in Figure~\ref{rnn_alignment}. We report the percentage of estimated gradient that points in the same direction as the true gradient. Our choice is motivated by the observation that as long as the approximate gradient estimate points in the same direction as the true gradient, we can expect the parameters to move in the right direction. We include T-BPTT as a baseline with truncation windows of 1, 3, 5, 20, and 40.  When lateral connections are few, the estimate of the gradient using \algoname{} is highly aligned with the true gradient. The misalignment increases as $s$ increases. This matches our expectations --- when $s$ is small, the approximation $\frac{\partial h_i(t)}{\theta_j}\approx0$ holds better. The gradient estimate is exact when there are no lateral connections. Sparsity has little effect on the gradient estimate of T-BPTT. The estimate of T-BPTT improves with the truncation window, but is worse than the estimate of \algoname{} for sparse \archi s. Even T-BPTT with a truncation window of 40 performs worse than \algoname{} when lateral connections are few. Note that the sequence length is only 50, and truncation window of 40 is very close to full BPTT. We repeated the experiment for a range of \archi s configurations and found the results to be consistent across different settings. For details of all the configurations, hyper-parameters, and results using the GRU cell, see Appendix~B.

\section{Credit-Assignment in Meta-learning}
Similar to recurrent state learning, meta-learning also requires structural credit assignment. Both BPTT and RTRL have been used successfully used for meta-learning. 

Finn~\etal~(2017) and Li~\etal~(2017) independently proposed using BPTT for learning through a stochastic gradient descent (SGD) update in a deep neural network. Javed and White~(2019) showed that T-BPTT can be used for meta-learning through long correlated sequences for learning without forgetting. 

RTRL has also been used for meta-learning. Sutton~(1992) showed that it can be used to learn the step-sizes for a linear predictor. Veeriah~\etal~(2017) extended Sutton's analysis by using it to approximate the meta-gradients for a one layer neural network. Applying RTRL to deep-learning based meta-learning methods is not tractable. MAML~(Finn~\etal~2017) --- a meta-learning algorithm that updates all parameters online at each learning step --- would require $O(|\theta|^2)$ memory and $O(|\theta|^3)$ operations per-step to recursively compute the meta-gradient. An alternative to MAML is the OML architecture --- independently proposed by Javed and White~(2019), and Bengio~\etal~(2020). The OML architecture updates only the parameters in the final prediction layers --- $W$ --- of the network at every step and updates the remaining parameters --- $\theta$ --- using the meta-gradient.  Computing meta-gradients for the OML architecture using RTRL requires $O(|W||\theta|)$ memory and $O(|W^2||\theta|)$ operations per-step. For $|W|<|\theta|$, this is already significantly more tractable than the $O(|\theta|^3)$ operations required by MAML. However, the memory and operations still do not scale linearly with the size of the network. 

We show that the OML architecture with a linear prediction layer can be combined with \archi s to approximate the meta-gradients online using $O(|\theta|)$ operations and memory. The key to our approximation, once again, is to exploit the modular structure of \archi s to remove insignificant gradient terms.

\subsection{Scalable Meta-learning using \archi s}

Once again we want to estimate $\frac{\partial L(t)}{\partial \theta_i}$, but with a key difference --- the network is updating the prediction parameters $w$ at every step. The parameters $w(t)$ linearly combine the state $h(t)$ for making predictions, as described in Equation~\ref{pln}. The weights $w_i(t)$ are updated as:

\begin{align}
    w_i(t+1) &= w_i(t) - \alpha_i \frac{\partial \mathcal L(t)}{\partial w_i(t)} \notag \\
    &= w_i(t) - \alpha_i \delta(t) \frac{\partial (y^*(t) - y(t)) }{\partial w_i(t)} \notag \\
    &= w_i(t) + \alpha_i \delta(t) \frac{\partial \sum_{j=1}^n w_j(t)h_j(t) }{\partial w_i(t)} \notag  \\
     &= w_i(t) + \alpha_i \delta(t) h_i(t) \label{lms}
\end{align}
where $a_i$ is the step-size parameter. Equation~\ref{lms} is the standard least mean squares (LMS) learning rule. The gradients for the loss $\mathcal L(t)$ w.r.t the parameters $\theta$ through the LMS learning rule can be computed as:

\begin{align}
          \frac{\partial \mathcal L(t)}{ \partial \theta_i} &= \frac{\partial \delta^2(t)}{ 2\partial \theta_i } \notag \\
          \frac{\partial \mathcal L(t)}{ \partial \theta_i} &= \delta(t)\frac{\partial \delta(t)}{ \partial \theta_i } \label{early} \\
          &= \delta(t)\frac{\partial(y^*(t) - \sum_{k=1}^{n} w_k(t) h_k(t))}{ \partial \theta_i}  \notag \\
                  &= - \delta(t)\sum_{k=1}^{n}  \frac{\partial (w_k(t) h_k(t))}{ \partial \theta_i} \notag \\
          & \text{$\frac{\partial w_k(t)}{\partial \theta_i}\ne0$ for an online learning network.} \notag\\
          & \text{Using product rule} \notag \\ 
           &= - \delta(t)\sum_{k=1}^{n} \left (  w_k(t)\frac{\partial  h_k(t)}{ \partial \theta_i} + h_k(t)\frac{\partial  w_k(t)}{ \partial \theta_i}\right) \label{before_meta}  \\
             & \text{Using the \algoname{} approximation, } \notag \\ 
             & \frac{\partial ( h_k(t)))}{ \partial \theta_i} \approx 0 \text{ when }i \ne k \implies  \notag \\
           &\approx - \delta(t)\left (  w_i(t)\frac{\partial  h_i(t)}{ \partial \theta_i} + \sum_{k=1}^{n}  h_k(t)\frac{\partial  w_k(t)}{ \partial \theta_i} \right) \label{after_meta} \\
           &\text{For \archi{} with sparse lateral connections,} \notag\\
           & \theta_i \text{ only indirectly influences } w_k \text{ when }i \ne k \notag \\ & \implies \frac{\partial ( w_k(t)))}{ \partial \theta_i}\approx 0  \implies  \label{crossprop} \\
           \frac{\partial \mathcal L(t)}{ \partial \theta_i} &\approx - \delta(t)\left (  w_i(t)\frac{\partial  h_i(t)}{ \partial \theta_i} +  h_i(t)\frac{\partial  w_i(t)}{ \partial \theta_i} \right) \label{after_approx}
    \end{align}
The approximation used in Equation~\ref{crossprop} is similar to the one proposed by Veeriah~\etal~(2017) for one layer networks. In deep networks, this is a poor approximation in the general case. We later show that the modular structure of \archi s is key for making this approximation work for multi-layer networks.
% It is also similar to the one used by Sutton~(1992) and Veeriah~\etal~(2017) for estimating meta-gradients for linear and one layer neural networks.

We can compute $\frac{\partial  h_i(t)}{ \partial \theta_i}$ in Equation~\ref{after_approx} in $O(|\theta|)$ as explained in the Sections~\ref{efficient_implementation}. $\frac{\partial  w_i(t)}{ \partial \theta_i}$ requires more work. Let: 

\begin{align}
 TW_{\theta_i}(t)  & \vcentcolon = \frac{\partial w_i(t)}{\partial \theta_i}  && \text{(By definition)} \label{defw_1}\\
 TW_{\theta_i}(0)  & \vcentcolon = 0  && \text{(By definition)} \label{defw_2} 
\end{align}

Then: 
\begin{align}
TW_{\theta_i}(t+1)   & = \frac{\partial w_i(t+1)}{\partial \theta_i}   \notag \\
& \text{Using LMS learning rule from Equation~\ref{lms}}  \notag \\
TW_{\theta_i}(t+1)   & = \frac{\partial (w_i(t) + \alpha_i \delta(t) h_i(t))}{\partial \theta_i} \notag  \\
& \text{Using Definition~\ref{defw_1}}  \notag \\
&= TW_{\theta_i}(t) +  \frac{\partial }{\partial \theta_i}(\alpha_i \delta(t) h_i(t)) \notag \\
& \text{Using product rule}  \notag \\
&= TW_{\theta_i}(t) +   \alpha_i \delta(t)  \frac{\partial h_i(t)}{\partial \theta_i} \notag \\ 
& + \alpha_i h_i(t)  \frac{\partial \delta(t)}{\partial \theta_i} \label{tw_update}
\end{align}

From Equation~\ref{early}~and~\ref{after_approx}, we can infer:
\begin{align}
    \frac{\partial \delta(t)}{\partial \theta_i} &\approx - w_i(t)\frac{\partial  h_i(t)}{ \partial \theta_i} - h_i(t)\frac{\partial  w_i(t)}{ \partial \theta_i} \notag \\
    &\approx - w_i(t)\frac{\partial  h_i(t)}{ \partial \theta_i} -  h_i(t) TW_i(t) \label{final_update}
\end{align}

All the terms in Equation~\ref{tw_update}~and~\ref{final_update} can be computed using $O(|\theta|)$ operations, allowing us to compute gradients through the LMS learning rule efficiently. Note that meta-learning requires an additional trace per parameter --- the $TW_\theta$ matrix.

\subsection{Making Sense of the Approximations}
\label{spec}
The meta-gradient computation introduces two additional approximations to the gradient. First, it simplifies Equation~\ref{before_meta} to Equation~\ref{after_meta} by assuming: 
\begin{equation*}
    \frac{\partial h_k(t)}{ \partial \theta_i}\approx 0 \text{ when }i \ne k
\end{equation*}
and second, it simplifies Equation~\ref{after_meta} to Equation~\ref{crossprop} by assuming: 
\begin{equation}
  \frac{\partial  w_k(t)}{ \partial \theta_i}\approx 0   \text{ when }i \ne k
\end{equation}
The first approximation is the same as one used to estimate the gradient through the recurrent states. We know that for \archi s with sparse lateral connections, it is a reasonable approximation. For the second approximation, we note from Equation~\ref{lms} that $\theta_i$ can influence the learning update to $w_k(t)$ by influencing two terms. First, it can influence $h_k(t)$ and second, it can influence the overall error $\delta(t)$. We know $\frac{\partial h_k(t)}{\partial \theta_i} \approx 0$ for \archi s with sparse lateral connections. Veeriah~\etal~(2017) and Sutton~(1992) empirically showed that the indirect effect of a parameter on the weight update through change in error is insignificant. This makes sense as $\frac{\partial \delta(t-1)}{\partial \theta_i} \frac{\partial w_k(t)}{\partial \delta(t-1)}$ is a product of two small numbers. 
\label{align_exp}
\begin{figure}
\centering
  \includegraphics[width=0.48\textwidth]{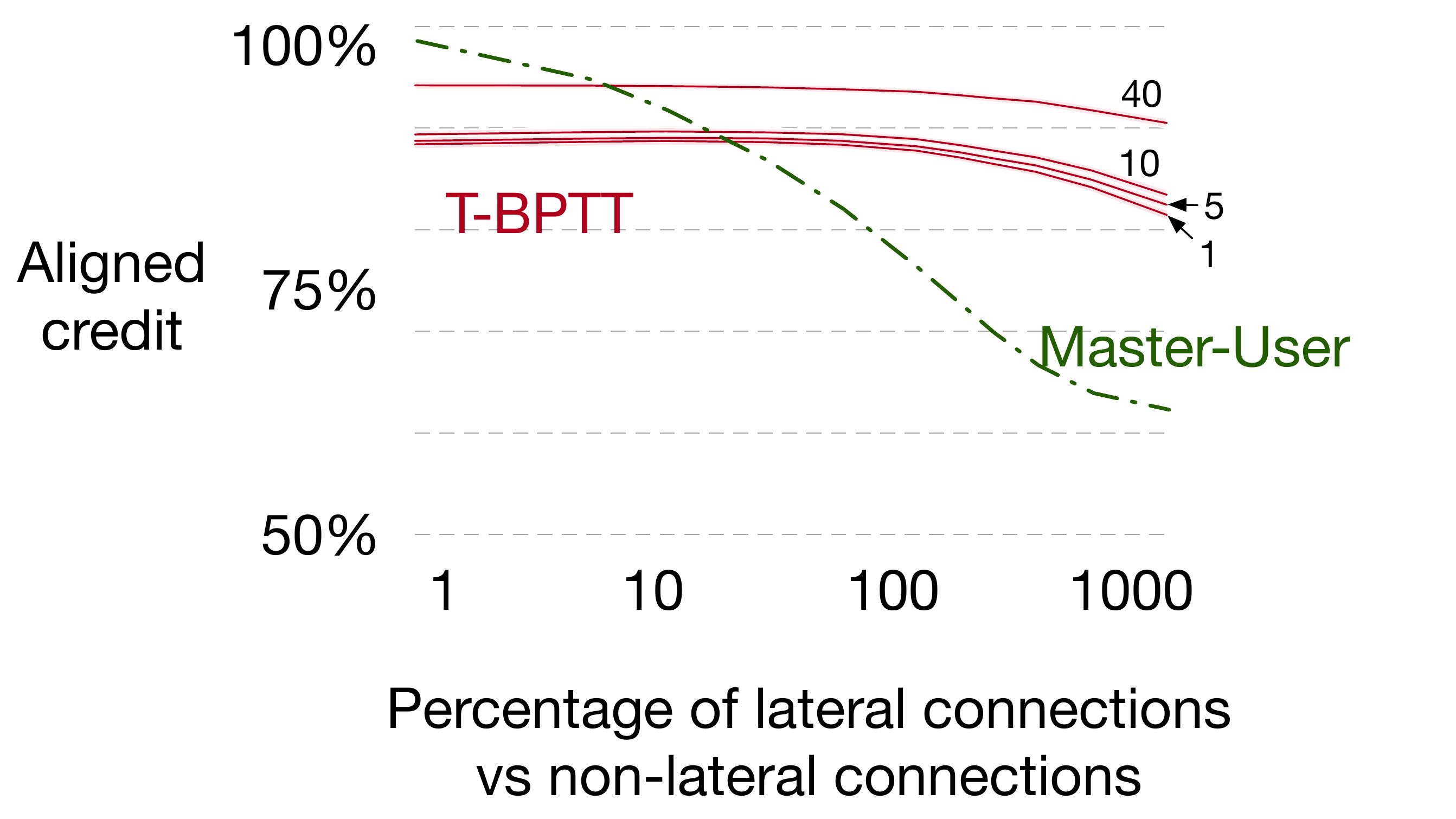}
  \caption{Quality of the approximation of the meta-gradient as a function of lateral connectivity. When number of lateral connections are low compared to non-lateral connections, the gradient estimate of \algoname{} for meta-learning is aligned with the true gradient. As lateral connections increase, the approximation gets worse. T-BPTT performs noticeably worse than \archi{} when lateral connections are sparse. The truncation window for T-BPTT is written at the end of each curve. All curves are averaged over 100 random seems, and the confidence intervals are one standard error.}
  \label{tbtt_meta}
\end{figure}

\begin{figure}[h]
\centering
  \includegraphics[width=0.48\textwidth]{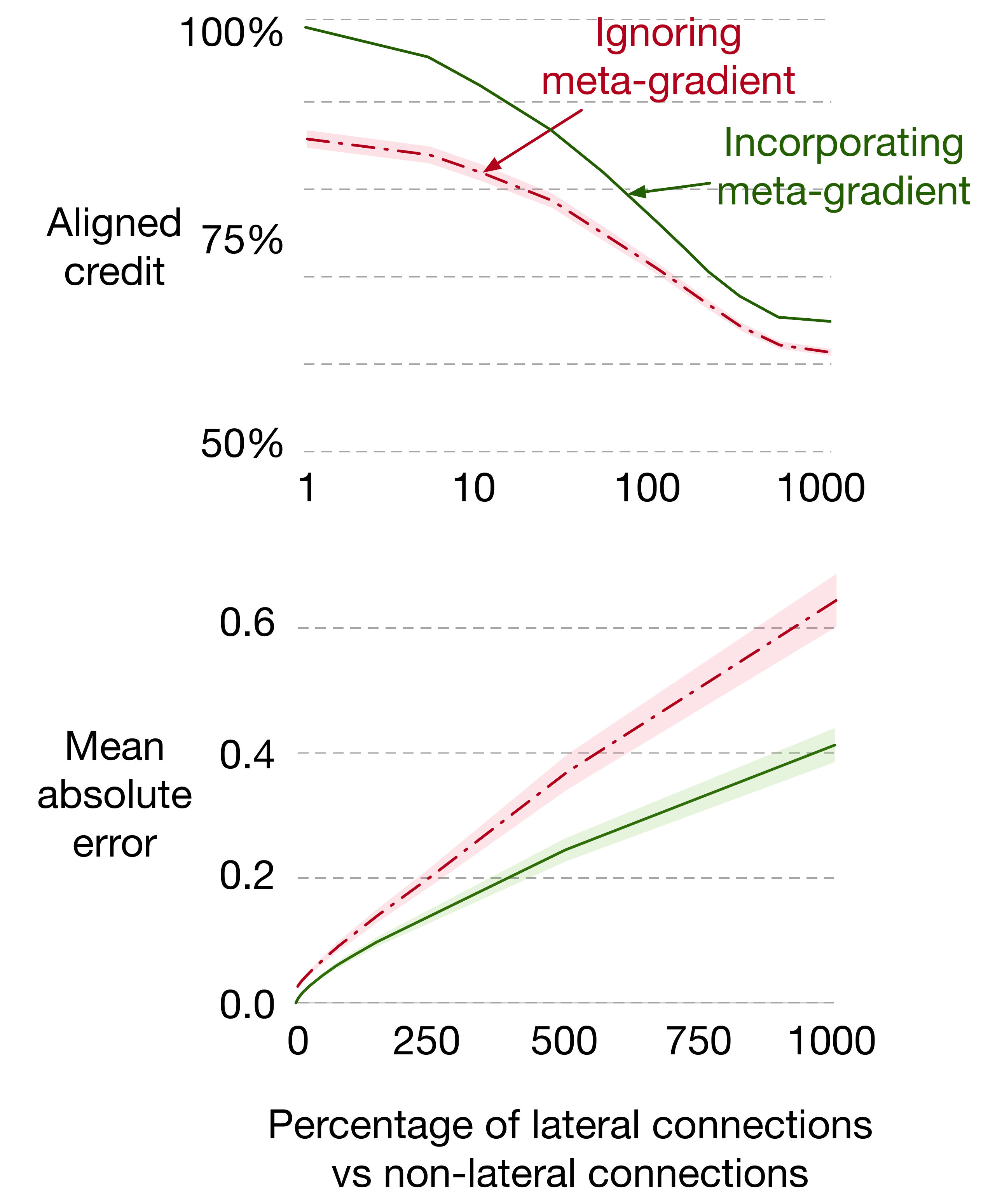}
  \caption{Ignoring the meta-gradient through the LMS update significantly impacts the accuracy of the estimate of the gradient. Even for a network with no lateral connections, less than 90\% of the gradients are unaligned when we ignore the meta-gradient. In many real-time recurrent systems, the last layer prediction weights are updated online and the meta-gradient is ignored. Our results suggest that doing so can be detrimental for effective credit-assignment.}
  \label{meta_alignment}
\end{figure}

\subsection{Evaluating Quality of the Gradient Estimate}
\subsubsection{Experiment Setup}
We use the test-bed described in Section~\ref{align_exp} to answer two questions. First, we evaluate how well $\algoname$ approximates the meta-gradient in isolation. We then test the importance of meta-gradients for estimating the true gradient of an online learning recurrent network. To answer the first question, we note that the error in the estimate of the meta-gradient can either be due to approximations made when computing gradients through the LMS rule or due to the approximations made when estimating gradient of the recurrent states. To disentangle the two sources of error, we modify the state update to be:
\begin{equation}
    h_i(t) = \sigma_g (\sum_{j=1}^n U^T_{ij} f_j(t)) \label{meta_used}
\end{equation}
to remove the recurrence. 
We use the same network architecture as Section~\ref{align_exp}. At every step, we update the weights $w$ using the LMS learning rule. We use a step-size of $1e^{-2}$ for each $w_i$ and compare the estimated gradient to the true gradient computed using BPTT. We use T-BPTT with truncation windows of 1, 5, 10, and 40 as baselines. For results for other choices of step-sizes and architectures, see Appendix~B.
\subsubsection{Results}
We report the results in Figure~\ref{tbtt_meta}.
\algoname{} performs better than T-BPTT with any truncation window when \archi s has sparse lateral connections. For a \archi{} with no lateral connections, the gradient estimate of \algoname{} is close to exact demonstrating that the effect of $\theta_i$ on $w_k(t)$ through error $\delta(t-1)$ is indeed insignificant as speculated in Section~\ref{spec}. We note that the benefit of \algoname{} for estimating the meta-gradient is less pronounced than the benefit for the recurrent gradient reported previously in Figure~\ref{rnn_alignment}. This could be either because for meta-gradients, long-term dependencies are not as important or because our test-bed does not capture the long-term dependencies that can otherwise emerge in meta-learning. Nonetheless, for up to 10\% ratio of lateral connection vs non-lateral connections, \algoname{} performs better than T-BPTT with a truncation window of 10 for estimating the meta-gradients.

Lastly, we investigate the importance of the meta-gradient to the overall gradient in an online learning recurrent network. We run \archi s on our test-bed twice, updating the final prediction weights $w$ at each step using the LMS rule with a step-size of $1e^{-2}$. In the first run, we estimate the gradient using Equation~\ref{after_approx}. This estimate takes into account gradient through both the recurrent state and the LMS weight update. In the second run, we ignore the gradient through the LMS weight update. This can be done by setting $\frac{\partial  w_i(t))}{ \partial \theta_i}$ to zero in Equation~\ref{after_approx}. We report the results in Figure~\ref{meta_alignment}. We find that ignoring the meta-gradient significantly impacts the accuracy of the overall gradient. Even for a columnar network with no lateral connections, less than $90\%$ of the gradients are aligned if we ignore the meta-gradient completely. We also report the average MAE between the estimates and the true gradient and find that ignoring the meta-gradient increases the error by $50\%$.

% \section{Connections to Neuroscience}

\section{Conclusion and Discussion} 
Scalable online recurrent learning is crucial for applying machine learning knowledge to real-world problems. In this work, we proposed one such method to achieve scalable online learning. Our method approximates the gradient for recurrent learning using $O(n)$ operations and memory. The approximations made by our method are interpretable. Additionally, the accuracy of the approximation can be controlled by controlling a tunable parameter. We also unify recurrent state learning and meta-learning and show that the same ideas can be applied to scale both. Finally, our method can be combined with arbitrary deep feature-extractors, opening the possibility of combining deep learning with online recurrent learning. There are key questions that our work leaves unanswered. Do modularized networks have limited capacity compared to fully connected networks? How does the accuracy of the gradient estimate changes as the network learn online? Can we increase the number of lateral connections overtime when features in columns have stabilized? We leave these questions for future work.

\nocite{sutton1992adapting}
\nocite{finn2017model}
\nocite{kingma2014adam}
\nocite{rumelhart1986learning}
\nocite{werbos1974beyond}
\nocite{werbos1988generalization}
\nocite{li2017meta}
\nocite{javed2019meta}
\nocite{bengio2019meta}
\nocite{williams1989learning}
\nocite{robinson1987utility}
\nocite{vivek}
\nocite{hochreiter1997long}
\nocite{menick2020practical}
\nocite{tallec2017unbiased}
\nocite{cooijmans2019variance}
\nocite{sutskever2013training}
\nocite{elman1990finding}
\nocite{mikolov2009neural}
\nocite{mikolov2010recurrent}
\nocite{ollivier2015training}
\nocite{bengio1990learning}
\nocite{cho2014learning}
\nocite{glorot2010understanding}
\nocite{mujika2018approximating}
\nocite{williams1990efficient}
\nocite{kapturowski2018recurrent}
\nocite{pllr}
\nocite{schmidhuber1987evolutionary}. 
\nocite{glorot2011deep}
\nocite{Tange2011a}
\bibliography{citations}

@inproceedings{finn2017model,
  title={Model-agnostic meta-learning for fast adaptation of deep networks},
  author={Finn, Chelsea and Abbeel, Pieter and Levine, Sergey},
  booktitle={ICML},
  year={2017},
}

@inproceedings{sutton1992adapting,
  title={Adapting bias by gradient descent: An incremental version of delta-bar-delta},
  author={Sutton, Richard S},
  booktitle={AAAI},
  year={1992},
  organization={San Jose, CA}
}

@article{kingma2014adam,
  title={Adam: A method for stochastic optimization},
  author={Kingma, Diederik P and Ba, Jimmy},
  journal={ICLR},
  year={2015}
}

@article{rumelhart1986learning,
	  title={Learning representations by back-propagating errors},
	  author={Rumelhart, David E and Hinton, Geoffrey E and Williams, Ronald J},
	  journal={Nature},
	  year={1986},
	  publisher={Nature Publishing Group}
	}

@article{werbos1974beyond,
	  title={Beyond regression:" new tools for prediction and analysis in the behavioral sciences},
	  author={Werbos, Paul},
	  journal={Ph. D. dissertation, Harvard University},
	  year={1974}
	}

@article{werbos1988generalization,
  title={Generalization of backpropagation with application to a recurrent gas market model},
  author={Werbos, Paul J},
  journal={Neural networks},
  year={1988},
  publisher={Elsevier}
}

@article{li2017meta,
  title={Meta-sgd: Learning to learn quickly for few-shot learning},
  author={Li, Zhenguo and Zhou, Fengwei and Chen, Fei and Li, Hang},
  journal={arXiv:1707.09835},
  year={2017}
}

@article{javed2019meta,
  title={Meta-learning representations for continual learning},
  author={Javed, Khurram and White, Martha},
  journal={NeurIPS},
  year={2019}
}

@article{bengio2019meta,
  title={A meta-transfer objective for learning to disentangle causal mechanisms},
  author={Bengio, Yoshua and Deleu, Tristan and Rahaman, Nasim and Ke, Rosemary and Lachapelle, S{\'e}bastien and Bilaniuk, Olexa and Goyal, Anirudh and Pal, Christopher},
  journal={ICLR},
  year={2020}
}

@book{robinson1987utility,
  title={The utility driven dynamic error propagation network},
  author={Robinson, AJ and Fallside, Frank},
  year={1987},
  publisher={University of Cambridge Department of Engineering Cambridge, MA}
}

@article{williams1989learning,
  title={A learning algorithm for continually running fully recurrent neural networks},
  author={Williams, Ronald J and Zipser, David},
  journal={Neural computation},
  year={1989}
}

@InProceedings{vivek,
author="Veeriah, Vivek
and Zhang, Shangtong
and Sutton, Richard S.",
title="Crossprop: Learning Representations by Stochastic Meta-Gradient Descent in Neural Networks",
booktitle="Machine Learning and Knowledge Discovery in Databases",
year="2017",
address="Cham"
}

@article{hochreiter1997long,
  title={Long short-term memory},
  author={Hochreiter, Sepp and Schmidhuber, J{\"u}rgen},
  journal={Neural computation},
  year={1997}
}

@article{menick2020practical,
  title={A Practical Sparse Approximation for Real Time Recurrent Learning},
  author={Menick, Jacob and Elsen, Erich and Evci, Utku and Osindero, Simon and Simonyan, Karen and Graves, Alex},
  journal={ICLR 2021},
  year={2020}
}

@article{tallec2017unbiased,
  title={Unbiased online recurrent optimization},
  author={Tallec, Corentin and Ollivier, Yann},
  journal={arXiv preprint arXiv:1702.05043},
  year={2017}
}

@article{mujika2018approximating,
  title={Approximating real-time recurrent learning with random kronecker factors},
  author={Mujika, Asier and Meier, Florian and Steger, Angelika},
  journal={Advances in Neural Information Processing Systems},
  year={2018}
}

@article{cooijmans2019variance,
  title={On the variance of unbiased online recurrent optimization},
  author={Cooijmans, Tim and Martens, James},
  journal={arXiv preprint arXiv:1902.02405},
  year={2019}
}

@book{sutskever2013training,
  title={Training recurrent neural networks},
  author={Sutskever, Ilya},
  year={2013},
  publisher={University of Toronto Toronto, Canada}
}

@article{elman1990finding,
  title={Finding structure in time},
  author={Elman, Jeffrey L},
  journal={Cognitive science},
  year={1990}
}

@inproceedings{mikolov2010recurrent,
  title={Recurrent neural network based language model},
  author={Mikolov, Tom{\'a}{\v{s}} and Karafi{\'a}t, Martin and Burget, Luk{\'a}{\v{s}} and {\v{C}}ernock{\`y}, Jan and Khudanpur, Sanjeev},
  booktitle={International speech communication association},
  year={2010}
}

@inproceedings{mikolov2009neural,
  title={Neural network based language models for highly inflective languages},
  author={Mikolov, Tomas and Kopecky, Jiri and Burget, Lukas and Glembek, Ondrej and others},
  booktitle={International conference on acoustics, speech and signal processing},
    year={2009}
}

@article{ollivier2015training,
  title={Training recurrent networks online without backtracking},
  author={Ollivier, Yann and Tallec, Corentin and Charpiat, Guillaume},
  journal={arXiv preprint arXiv:1507.07680},
  year={2015}
}

@techreport{bengio1990learning,
  title={Learning a synaptic learning rule},
  author={Bengio, Yoshua and Bengio, Samy and Cloutier, Jocelyn},
  year={1990},
}

@inproceedings{kapturowski2018recurrent,
  title={Recurrent experience replay in distributed reinforcement learning},
  author={Kapturowski, Steven and Ostrovski, Georg and Quan, John and Munos, Remi and Dabney, Will},
  booktitle={International conference on learning representations},
  year={2018}
}

@article{cho2014learning,
  title={Learning phrase representations using RNN encoder-decoder for statistical machine translation},
  author={Cho, Kyunghyun and Van Merri{\"e}nboer, Bart and Gulcehre, Caglar and Bahdanau, Dzmitry and Bougares, Fethi and Schwenk, Holger and Bengio, Yoshua},
  journal={arXiv preprint arXiv:1406.1078},
  year={2014}
}

@inproceedings{glorot2010understanding,
  title={Understanding the difficulty of training deep feedforward neural networks},
  author={Glorot, Xavier and Bengio, Yoshua},
  booktitle={International conference on artificial intelligence and statistics},
  year={2010},
  organization={JMLR Workshop and Conference Proceedings}
}

@article{williams1990efficient,
  title={An efficient gradient-based algorithm for on-line training of recurrent network trajectories},
  author={Williams, Ronald J and Peng, Jing},
  journal={Neural computation},
  year={1990},
}

@InProceedings{pllr,
  title = 	 {Meta-Learning with Memory-Augmented Neural Networks},
  author = 	 {Adam Santoro and Sergey Bartunov and Matthew Botvinick and Daan Wierstra and Timothy Lillicrap},
  booktitle = 	 {International Conference on Machine Learning},
    year = 	 {2016}
}

@phdthesis{schmidhuber1987evolutionary,
  title={Evolutionary principles in self-referential learning, or on learning how to learn: the meta-meta-... hook},
  author={Schmidhuber, J{\"u}rgen},
  year={1987},
  school={Technische Universit{\"a}t M{\"u}nchen}
}

@inproceedings{glorot2011deep,
  title={Deep sparse rectifier neural networks},
  author={Glorot, Xavier and Bordes, Antoine and Bengio, Yoshua},
  booktitle={Proceedings of the fourteenth international conference on artificial intelligence and statistics},
  year={2011},
  organization={JMLR Workshop and Conference Proceedings}
}

@article{Tange2011a,
  title = {GNU Parallel - The Command-Line Power Tool},
  author = {O. Tange},
  address = {Frederiksberg, Denmark},
  journal = {;login: The USENIX Magazine},
  month = {Feb},
  number = {1},
  volume = {36},
  url = {http://www.gnu.org/s/parallel},
  year = {2011},
  pages = {42-47},
  doi = {http://dx.doi.org/10.5281/zenodo.16303}
}

\bibliographystyle{icml2021}

\onecolumn 
\appendix
\section{Information Decay in Randomly Initialized LSTM and GRU Networks}
In the main text, we reported results using an RNN with recurrence defined by Equation~\ref{recurrent_used}. Our recurrence relationship simply adds to the hidden state to get the new state. This results in a challenging benchmark for gradient-estimation in which parameter can influence a target many steps in the future. A randomly initialized LSTM or GRU cell, on the other hand, is not guaranteed to have long-range dependencies. To show this empirically, we measure how many steps can information propagate in untrained LSTM and GRU networks. We set the internal states of the networks to be a vector of one, and update the state for 20 steps using a zero vector as input. Using a zero input vector prevents addition of any new information in the state, and allows us to measure the decay of initial value of the state. 

 \begin{figure*}
\centering
  \includegraphics[width=\textwidth]{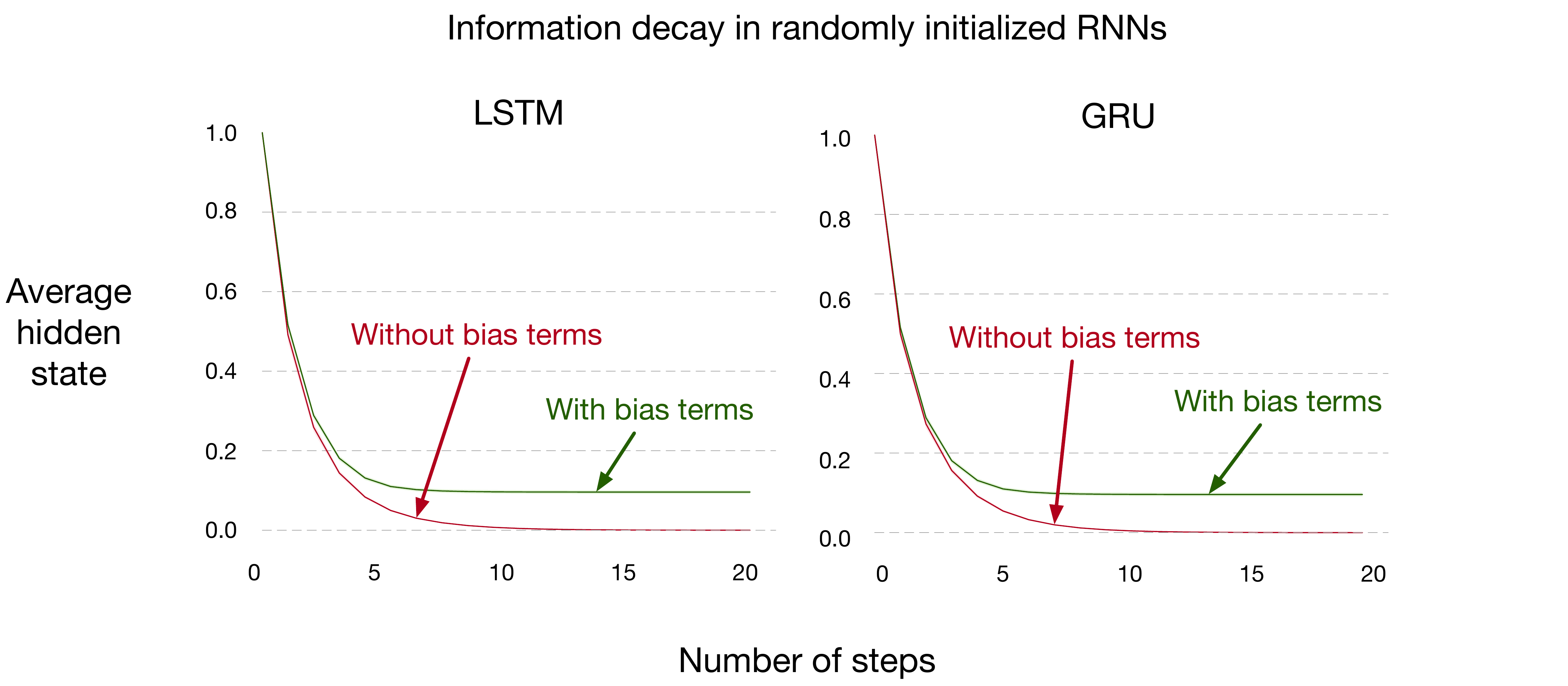}
  \caption{Information decay in randomly initialized LSTM (left) and GRU (right) cells. We initialize the hidden state for both to be 1, and then run each for 20 steps. Both networks see a zero vector as input. The hidden state for both LSTM and GRU without bias terms decays to zero quickly showing that these architectures, when untrained, do not involve long-range dependencies. This makes them a poor candidate for comparing gradient approximation methods. When the bias terms are not zero, the hidden state converges to a non-zero constant due to the bias term. }
  \label{decay_lstm}
\end{figure*}

 \begin{figure*}
\centering
  \includegraphics[width=0.9\textwidth]{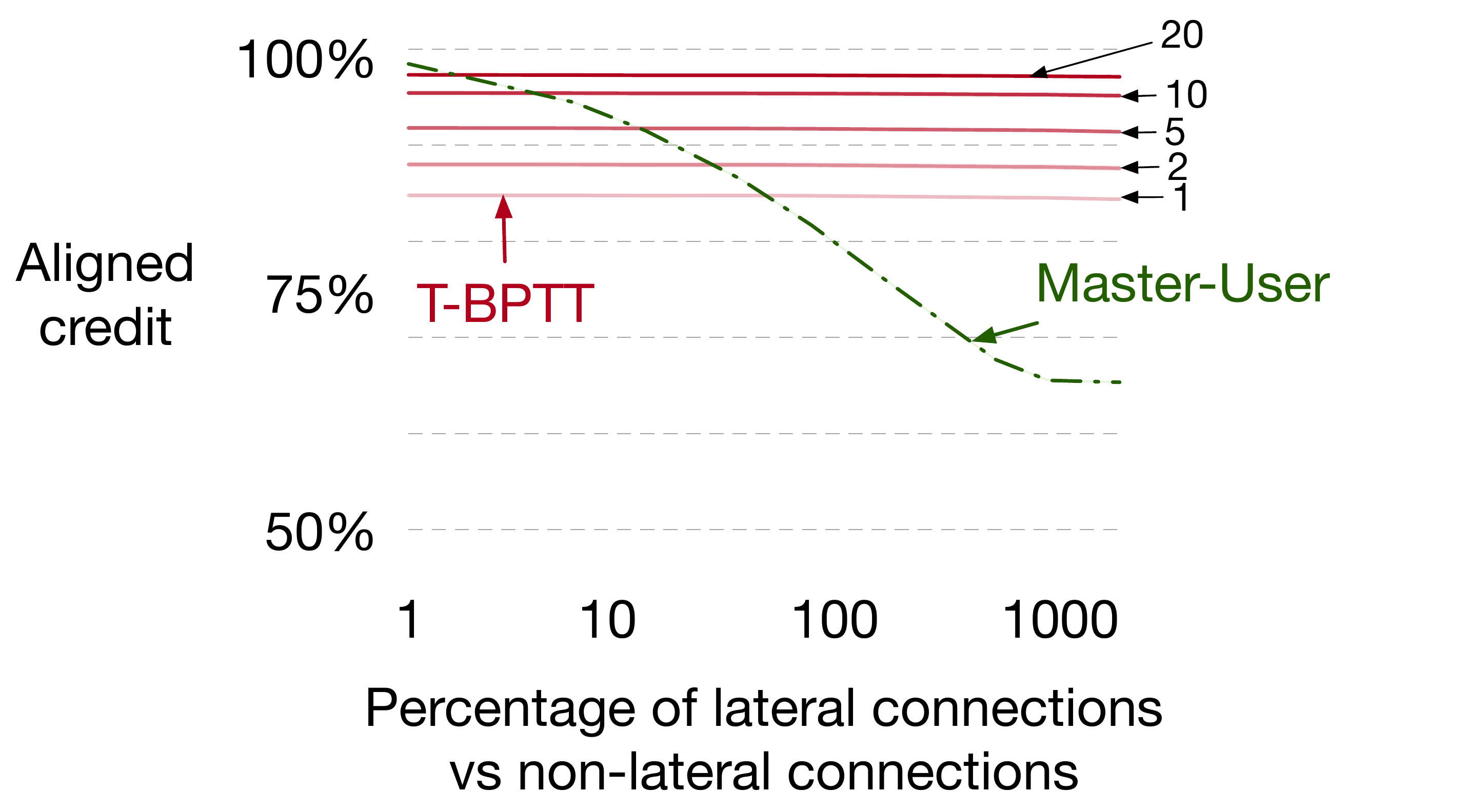}
  \caption{Quality of the approximation as a function of lateral connectivity for a GRU based RNN. The value of truncation window for T-BPTT is indicated by numbers at the end of each curve. A randomly initialized GRU decays information in the hidden state rapidly. This improves the performance of T-BPTT with small truncation windows. However, the good performance of T-BPTT is expected to deteriorate once the GRU call has learned longer-range dependencies. }
  \label{gru_results}
\end{figure*}

 \begin{figure}[h]
\centering
  \includegraphics[width=\textwidth]{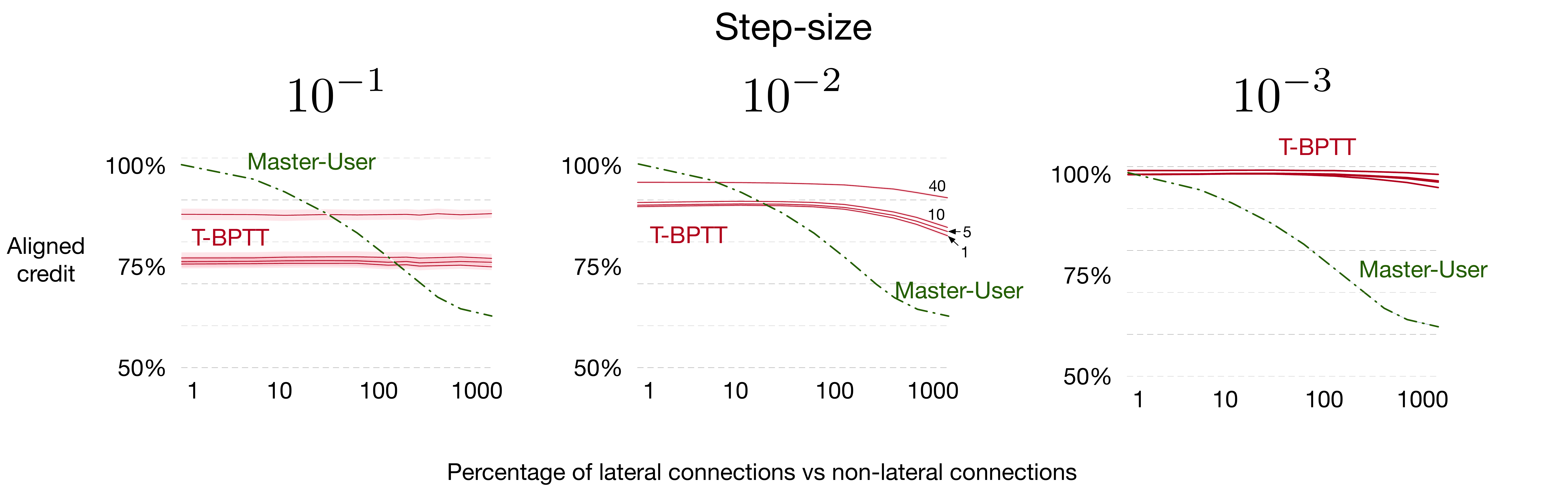}
  \caption{Effect of step-size parameter on gradient approximation for T-BPTT and Master-User. When step-size is large, T-BPTT is a poor approximation to the true gradient. Even a truncation window of 40 --- almost equal to the sequence length 50 --- is not sufficient to approximate the meta-gradient well. The error introduced by Master-User, on the other hand, does not vary with step-size. This makes Master-User a promising algorithm for meta-learning with large step-sizes.}
  \label{meta_lr_eff}
\end{figure}

We initialize the LSTM and GRU parameters by uniformly sampling the weights between $(-\sqrt{k} , \sqrt{k}) $ where $k$ is $\frac{1}{\textit{hidden units}}$. This is the default initialization used by Pytorch~(Paszke\etal, 2019). We report the results for LSTM and GRU networks with and without the bias terms for $k=50$ in Figure~\ref{decay_lstm}. Other values of $k$ give similar results. From Figure~\ref{decay_lstm}, we see that the information in LSTM and GRU cells decays rapidly. Without any bias terms, the hidden state is close to zero with-in just 5 steps and approaches zero in 10 steps. This suggests that for randomly initialized GRU and LSTM cells, T-BPTT with a small truncation window would perform well. 

The rapid decay in LSTM and GRU cells is not surprising. The value of the forget gate for untrained LSTM and GRU cells initialized using popular initialization methods is $\approx 0.5$ which makes information decay exponentially. Note that trained LSTM and GRU cells do not suffer from this --- their forget gate can output 1 resulting in no decay of information.

\subsection{Results with GRU Networks} 
We run experiment described in Section~\ref{align_exp} using randomly initialized GRU cells and report the results in Figure~\ref{gru_results}. Because randomly initialized GRUs decay information quickly, T-BPTT with small truncation window performs significantly better than it does in Figure~\ref{rnn_alignment}. Note that the good performance of T-BPTT can only be expected to hold on the untrained GRU RNN. Once the GRU cell starts learning, the decay rate of information can reduce --- depending on the learning signal --- and the gradient approximation of T-BPTT could deteriorate. \archi s trained with \algoname{} do no suffer from this --- their error in gradient-estimation stays low for sparse lateral connections even when there are long-range dependencies, as shown in Figure~\ref{rnn_alignment}.

\section{Experiment details} 
\subsection{Hyper-parameters}

The key parameters in our experiments are number of columns (C), number of units in each layer of the column (W), and the step-size (SS). We tried C = 5, 10, 20, 50, 100; W = 5, 10, 20, 50, 100; and SS = $10^{-1}, 10^{-2}, 10^{-3}$. SS is a parameter only for the meta-learning experiment. C and W result in 25 combinations for the RNN experiments and wherease C, W, and SS result in 75 combinations for the meta-learning experiments. We did 100 runs for each configuration resulting in a total of 10,000 runs. We found that changing C and W does not change the results qualitatively and the results are similar to those reported in Figure~\ref{rnn_alignment}. The SS parameter does influence the result noticeably and we report those in the next section. 

\subsection{Effect of Step-size on Approximation of the Meta-gradient}
We report the effect of changing SS for a network with C = 20 and W = 50 in Figure~\ref{meta_lr_eff}. When SS is large, T-BPTT performs worse. This makes sense --- a large step-size increases the influence of parameters on future predictions and T-BPTT with small truncation window is a poor approximation to the gradient. For smaller step-sizes, the effect of parameters on future predictions is minuscule and even T-BPTT with a truncation window of 1 is a good approximation. The error in approximation due to Master-User is independent of the step-size. This makes Master-User ideal for meta-learning with large step-sizes.

\subsection{Compute Resources and Reproducibility}
We used two 48 core CPU servers for all the experiments. All experiments in the paper can be reproduced in less than 5 hours using the two servers. We ran each configuration using 100 random seeds. Running each configuration only once can be done on a personal computer in a few hours. We provide anonymized code for reproducing the experiments at \url{https://github.com/khurramjaved96/columnar_networks}. 

\section*{References}
Paszke, Adam, et al. "Pytorch: An imperative style, high-performance deep learning library." arXiv preprint arXiv:1912.01703 (2019).

%%%%%%%%%%%%%%%%%%%%%%%%%%%%%

\end{document}